\documentclass[letterpaper]{article} % DO NOT CHANGE THIS
\usepackage{aaai24}  % DO NOT CHANGE THIS
\usepackage{times}  % DO NOT CHANGE THIS
\usepackage{helvet}  % DO NOT CHANGE THIS
\usepackage{courier}  % DO NOT CHANGE THIS
\usepackage[hyphens]{url}  % DO NOT CHANGE THIS
\usepackage{graphicx} % DO NOT CHANGE THIS
\usepackage{natbib}  % DO NOT CHANGE THIS AND DO NOT ADD ANY OPTIONS TO IT
\usepackage{caption} % DO NOT CHANGE THIS AND DO NOT ADD ANY OPTIONS TO IT
\usepackage{algorithm}
\usepackage{algorithmic}

\usepackage{newfloat}
\usepackage{listings}
\DeclareCaptionStyle{ruled}{labelfont=normalfont,labelsep=colon,strut=off} % DO NOT CHANGE THIS
\floatstyle{ruled}
\newfloat{listing}{tb}{lst}{}
\floatname{listing}{Listing}
\usepackage{tikz}
\usepackage{multirow}
\usepackage{colortbl}
\usepackage{booktabs}
\usepackage{adjustbox}
\usepackage{subcaption}

\title{Cause and Effect: Can Large Language Models Truly Understand Causality?}
\author {
    Swagata Ashwani\textsuperscript{\rm 1},
    Kshiteesh Hegde\textsuperscript{\rm 2},
    Nishith Reddy Mannuru\textsuperscript{\rm 3},
    Dushyant Singh Sengar\textsuperscript{\rm 4},
    Mayank Jindal\textsuperscript{\rm 4},
    Krishna Chaitanya Rao Kathala\textsuperscript{\rm 5},
    Dishant Banga\textsuperscript{\rm 6},
    Vinija Jain\textsuperscript{\rm 7},
    Aman Chadha\textsuperscript{\rm 7,8}\thanks{Work done outside position at Amazon.}
}
\affiliations {
    \textsuperscript{\rm 1}Carnegie Mellon University\\ sashwani@alumni.cmu.edu\\
    \textsuperscript{\rm 2}Rensselaer Polytechnic Institute,\\
    \textsuperscript{\rm 3}University of North Texas,\\
    \textsuperscript{\rm 4}Independent Researcher,\\
    \textsuperscript{\rm 5}University of Massachusetts,\\
    \textsuperscript{\rm 6}Bridgetree,\\
    \textsuperscript{\rm 7}Stanford University, \\
        \textsuperscript{\rm 8}Amazon GenAI\\
    
}
\begin{document}

\maketitle

\usetikzlibrary{shapes, arrows, positioning}

\tikzset{
    premise/.style={
           rectangle,
           rounded corners,
           draw=black, very thick,
           fill=blue!20,
           text width=7em, % Adjusted width
           minimum height=2em,
           text centered,
           scale=0.8, % Scaling down the node sizes
    },
    hypothesis/.style={
           rectangle,
           rounded corners,
           draw=black, very thick,
           fill=green!20,
           text width=7em, % Adjusted width
           minimum height=2em,
           text centered,
           scale=0.8, % Scaling down the node sizes
    },
    correct/.style={
           rectangle,
           rounded corners,
           draw=black, very thick,
           fill=yellow!20,
           text width=7em, % Adjusted width
           minimum height=2em,
           text centered,
           scale=0.8, % Scaling down the node sizes
    },
    pil/.style={
           ->,
           thick,
           shorten <=2pt,
           shorten >=2pt,
    },
}

\begin{abstract}
With the rise of Large Language Models (LLMs), it has become crucial to understand their capabilities and limitations in deciphering and explaining the complex web of causal relationships that language entails. Current methods use either explicit or implicit causal reasoning, yet there is a strong need for a unified approach combining both to tackle a wide array of causal relationships more effectively. This research proposes a novel architecture called Context-Aware Reasoning Enhancement with Counterfactual Analysis (CARE-CA) to enhance causal reasoning and explainability. The proposed framework incorporates an explicit causal detection module with ConceptNet and counterfactual statements, as well as implicit causal detection through LLMs. Our framework goes one step further with a layer of counterfactual explanations to accentuate LLMs' understanding of causality. The knowledge from ConceptNet enhances the performance of multiple causal reasoning tasks such as causal discovery, causal identification, and counterfactual reasoning. The counterfactual sentences add explicit knowledge of `not caused by' scenarios. By combining these powerful modules, our model aims to provide a deeper understanding of causal relationships, enabling enhanced interpretability. 
Evaluation of benchmark datasets shows improved performance across all metrics, such as accuracy, precision, recall, and F1 scores.
We also present CausalNet, a novel dataset specifically curated to benchmark and enhance the causal reasoning capabilities of LLMs. This dataset is accompanied by code designed to facilitate further research in this domain.
\end{abstract}

\section{Introduction}

As Large Language Models (LLMs) play an increasingly central role in technology, their ability to understand and logically navigate causal relationships becomes essential since they impact the trust their users have on them. \cite{kiciman2023causal} This skill is paramount for refining the depth and applicability of LLMs in complex scenarios, driving advancements that hinge on nuanced interpretations of cause and effect.

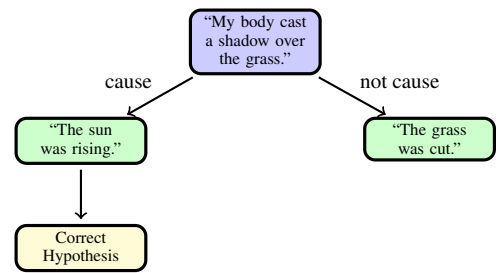
\begin{figure}[!ht]
\centering
\begin{tikzpicture}[node distance=1cm, auto, scale=0.8, transform shape]
  % Nodes
  \node[premise] (premise) {``My body cast a shadow over the grass."};
  \node[hypothesis, below left=of premise] (hyp1) {``The sun was rising."};
  \node[hypothesis, below right=of premise] (hyp2) {``The grass was cut."};
  \node[correct, below=of hyp1] (correct) {Correct Hypothesis};
  
  % Edges
  \draw[pil] (premise) -- node[above left] {cause} (hyp1);
  \draw[pil] (premise) -- node[above right] {not cause} (hyp2);
  \draw[pil] (hyp1) -- (correct);
  
\end{tikzpicture}
\caption{Causal reasoning without CARE-CA: Given the premise ``My body cast a shadow over the grass.", the left hypothesis, ``The sun was rising," should be identified as the cause to arrive at the correct hypothesis conclusion.}
\label{fig:first_example}
\end{figure}

\usetikzlibrary{shapes, arrows, positioning, fit, calc, backgrounds}

\tikzset{
    block/.style={
           rectangle,
           rounded corners,
           draw=black, very thick,
           fill=blue!20,
           text width=10em,
           minimum height=2em,
           text centered,
           scale=0.8,
    },
    line/.style={
           draw, thick, ->, shorten >=2pt,
    },
    decision1/.style={
           diamond,
           draw,
           text width=4.5em,
           text badly centered,
           fill=green!20,
           inner sep=0pt
    },
        decision2/.style={
           diamond,
           draw,
           text width=4.5em,
           text badly centered,
           fill=yellow!20,
           inner sep=0pt
    },
}

\begin{figure}[!ht]
\centering

\begin{tikzpicture}[auto, scale=0.8, transform shape]
  % Nodes
  \node[block] (premise) {``My body cast a shadow over the grass."};
  \node[decision1, below left=2cm and 1cm of premise] (hyp1) {``The sun was rising."};
  \node[decision2, below right=2cm and 1cm of premise] (hyp2) {``The grass was cut."};
  \node[block, below=6cm of premise] (correct) {Correct Hypothesis};
  
  \node[block, below=of correct] (conceptnet) {``ConceptNet Integration: `Shadows' related to `light source'.''};
  \node[block, below=of conceptnet] (contextual) {``Contextual Prompting: Hypotheses contextualized with time of day.''};
  \node[block, below=of contextual] (counterfactual) {``Counterfactual Reasoning: `What if no light source?' scenario.''};
  \node[block, below=of counterfactual] (improvement) {``Improved Causal Reasoning: Correct hypothesis identified with context and counterfactuals.''};

  % Edges
  \draw[line] (premise) -| (hyp1);
  \draw[line] (premise) -| (hyp2);
  \draw[line] (hyp1) -- node[left] {cause} (correct);
  \draw[line] (hyp2) -- node[right] {not cause} (correct);
  \draw[line] (correct) -- (conceptnet);
  \draw[line] (conceptnet) -- (contextual);
  \draw[line] (contextual) -- (counterfactual);
  \draw[line] (counterfactual) -- (improvement);
  
\end{tikzpicture}
\caption{Causal Reasoning Enhanced with CARE-CA: Starting from a premise, causal hypotheses are evaluated. Integration of external knowledge from ConceptNet enhances understanding. Contextual prompting adapts hypotheses to the time of day. Counterfactual reasoning explores alternative scenarios. Improved causal reasoning is achieved by incorporating context and counterfactuals, leading to the identification of the correct hypothesis.}
\label{fig:second_example}
\end{figure}
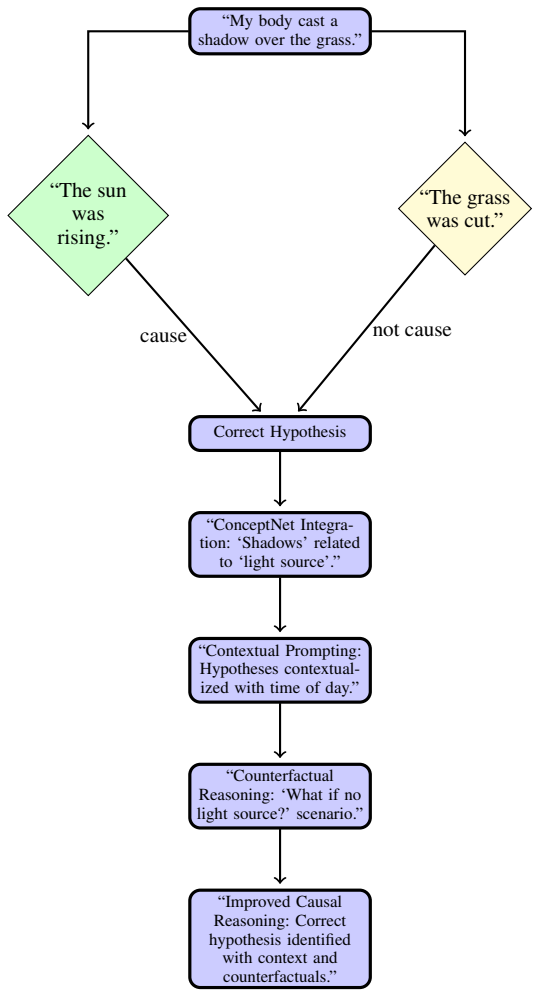

Given the growing reliance on AI systems to make consequential, mission-critical decisions, we need to enhance the causal reasoning capabilities of LLMs. Prior research \cite{weng2023large,zhang2023understanding} has revealed significant limitations in LLMs' causal reasoning capabilities. While they may mimic causal language, most need a genuine comprehension of causal mechanisms. This is concerning as it could propagate misinformation or lead to unreliable predictions. Bridging this causal reasoning gap is an active area of research.

Enhancing the causal reasoning capabilities of LLMs can significantly impact their reliability and trustworthiness across many applications. A more robust causal understanding of LLMs could improve healthcare and public policy decision-making \cite{pena2023leveraging}. It also promises to enhance interpretability and transparency.

However, prevailing approaches need help with flexibility and depth of causal inference. This work investigates whether these advanced models, like BERT \cite{devlin2018bert}, RoBERTa \cite{liu2019roberta}, XLM-RoBERTa \cite{conneau2019unsupervised}, ALBERT \cite{lan2019albert}, DeBERTa \cite{he2020deberta}, Llama 2 \cite{touvron2023llama}, T5 \cite{raffel2020exploring}, Mistral \cite{jiang2023mistral}, GPT-3.5 \cite{gpt3.5}, and Gemini Pro \cite{team2023gemini}, can truly grasp and articulate causal relationships, a cornerstone in the journey towards Artificial General Intelligence (AGI). We explore this through a blend of theoretical analysis and empirical investigation, focusing on the capability of LLMs to comprehend and articulate causality in the literal sense.

Building on this foundation, we introduce the CARE-CA framework, a novel architecture designed to amplify the causal reasoning competence of LLMs. The CARE-CA framework is distinct in its use of explicit knowledge integration from resources like ConceptNet \cite{speer2017conceptnet} and implicit reasoning patterns derived from models such as BERT. This dual approach bridges the gap between knowledge-driven and data-driven inference. It enhances the model's performance across four critical domains of causal reasoning: Causal Relationship Identification, Causal Discovery, Causal Explanation, and Counterfactual Reasoning.

We present a comprehensive suite of evaluation metrics, including accuracy, F1, precision, recall, and  human evaluation, to assess and compare the performance of existing LLMs against our proposed CARE-CA framework. Furthermore, we introduce a new dataset, CasualNet, which, we experimentally demonstrate, boosts LLMs' causal reasoning ability. CasualNet is poised to serve as a benchmark for future advancements in this field, providing a rigorous testing ground for emerging AI models.

By uniting explicit and implicit causal modules alongside contextual and counterfactual enhancements, this research nudges LLMs towards improved causal reasoning — a pivotal step in unraveling AI’s black box and realizing more trustworthy, explainable systems.

\section{Related Work}
Prior research has explored various approaches to understand and enhance causal reasoning capabilities of LLMs. There have been claims that LLMs can only mimic causal language and they lack genuine causal understanding, calling them ``causal parrots'' \cite{zevcevic2023causal}. So, assessing the ability of LLMs to answer causal questions, discussing their strengths and weaknesses is vital. To this end, we further explore the potential of integrating explicit and implicit causal modules to improve LLM performance \cite{zhang2023understanding}, This is a key principle underlying our CARE-CA framework.

One remarkable work is the CRAB benchmark \cite{romanou2023crab}, which evaluates the ability of LLMs to infer causal relationships between real-world events. The authors found that while LLMs can perform well on certain causal reasoning tasks, they struggle with more complex scenarios that require a deeper understanding of causality.

Another work showed that LLMs can infer causation from correlation, a crucial skill for causal reasoning \cite{jin2023can}. Their findings suggest that while LLMs can learn some causal patterns, they often fail to distinguish between causal and non-causal relationships, highlighting the need for more targeted approaches. 

Additionally, \cite{jin2021causal} explored the impact of the causal direction of data collection on the performance of LLMs in causal reasoning tasks. They found that models trained on data with a specific causal direction perform better on tasks that align with that direction, underscoring the importance of dataset design in causal reasoning research. These studies provide a solid foundation for understanding the current state of causal reasoning in LLMs. 

Given the widespread implications of LLM causal reasoning capabilities, we aim to enhance the effectiveness of all four aspects of causal reasoning in addition to the LLM evaluation work done before \cite{zhuang2023lens}. Our method will specifically focus on enhancing the causal reasoning by incorporating explicit knowledge from knowledge graphs such as ConceptNet.

While various past works have demonstrated the superior performance of GPT-3.5 and Gemini Pro in certain causal reasoning tasks, their work did not provide a concrete architecture to enhance these capabilities. In contrast, our CARE-CA framework goes a step further by proposing a novel hybrid approach that combines explicit causal knowledge from resources like ConceptNet introduced by \cite{speer2017conceptnet} with the implicit reasoning capabilities of LLMs. 

\textbf{Contributions:} CARE-CA aims to provide a more comprehensive and effective solution for tackling a wider array of causal reasoning tasks by incorporating counterfactual reasoning and contextual prompting. Unlike previous methods that either relied on explicit or implicit causal reasoning, CARE-CA's unique integration of these two complementary approaches sets it apart, allowing for a more robust and flexible causal understanding. This distinction enables CARE-CA to potentially outperform existing techniques in tasks such as causal relationship identification, counterfactual reasoning, and causal discovery, as demonstrated in our experimental evaluation.

Furthermore, our methodological advancements are showcased through the development and utilization of the CausalNet dataset, specifically designed to benchmark and refine the causal reasoning capabilities of LLMs. By focusing on the four key aspects of causal reasoning—Causal Relationship Identification, Counterfactual Reasoning, Causal Discovery, and Causal Explanation—CARE-CA represents a comprehensive approach to enhancing LLMs' causal reasoning faculties.

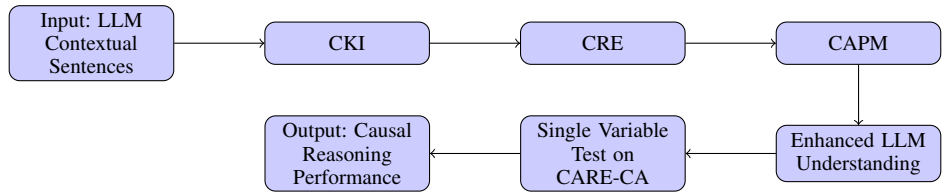
\begin{figure*}[!ht]
\centering
\begin{tikzpicture}[
  auto,
  node distance=1cm and 1.5cm,
  scale=0.8,
  transform shape,
  align=center,
  block/.style={rectangle, draw, text width=2.5cm, text centered, rounded corners, minimum height=2em, fill=blue!20}
]
  % Nodes
  \node[block] (conceptnet) {Input: LLM\\Contextual\\Sentences};
  \node[block, right=of conceptnet] (contextual) {CKI};
  \node[block, right=of contextual] (counterfactual) {CRE};
  \node[block, right=of counterfactual] (capm) {CAPM};
  \node[block, below=of capm] (enhanced) {Enhanced LLM\\Understanding};
  \node[block, left=of enhanced] (singlevar) {Single Variable\\Test on\\CARE-CA};
  \node[block, left=of singlevar] (causal) {Output: Causal\\Reasoning\\Performance};

  % Edges
  \draw[->] (conceptnet) -- (contextual);
  \draw[->] (contextual) -- (counterfactual);
  \draw[->] (counterfactual) -- (capm);
  \draw[->] (capm) -- (enhanced);
  \draw[->] (enhanced) -- (singlevar);
  \draw[->] (singlevar) -- (causal);
\end{tikzpicture}
% \begin{tikzpicture}[auto, scale=0.8, transform shape]
%   % Nodes
%   % \node[block] (start) {Start};
%   \node[block] (conceptnet) {Input: LLM Contextual Sentences.};
%   \node[block, right=of conceptnet, minimum width=0.2cm] (contextual) {CKI};
%   \node[block, right=of contextual] (counterfactual) {CRE};
%   \node[block, right=of counterfactual] (capm) {CAPM};
%   \node[block, right=of capm] (enhanced) {Enhanced LLM Understanding};
%   \node[block, right=of enhanced] (singlevar) {Single Variable Test on CARE-CA};
%   \node[block, right=of singlevar] (causal) {Output: Causal Reasoning Performance};
%   % \node[block, right=of causal] (end) {End};

%   % Edges
%   % \draw[line] (start) -- (conceptnet);
%   \draw[line] (conceptnet) -- (contextual);
%   \draw[line] (contextual) -- (counterfactual);
%   \draw[line] (counterfactual) -- (capm);
%   \draw[line] (capm) -- (enhanced);
%   \draw[line] (enhanced) -- (singlevar);
%   \draw[line] (singlevar) -- (causal);
%   % \draw[line] (causal) -- (end);

% \end{tikzpicture}
\caption{Enhancing LLM Causal Understanding via Structured Knowledge and Counterfactuals: This approach integrates ConceptNet knowledge graphs and 'what-if' scenarios to improve LLMs' causal reasoning, using CKI, CRE, and CAPM to boost performance on causal benchmarks like CARE-CA.}
\label{fig:third_example}
\end{figure*}

\section{Approach}
Our approach combines the explicit, structured causal reasoning of ConceptNet knowledge graphs coupled with counterfactual sentences to improvise the causal understanding of LLMs. This novel architecture aims to surpass traditional decoder or encoder-only models by leveraging the rich semantic knowledge base of ConceptNet with advanced contextual inference capabilities and `alternate scenarios' of the contextual sentences to further aid the LLMs in understanding the causality of scenarios. The combination of the above provides relevant contextual information for the LLMs to understand the causal reasoning in question.  We carry out a single variable test comparing the performance (X and Y) on CARE-CA using accuracy, recall, precision and F1 scores.

We illustrate the components of CARE-CA in Figure \ref{fig:third_example} and expand on the critical components briefly.

1. \textbf{Contextual Knowledge Integrator (CKI)} enriches the AI's reasoning process with relevant external knowledge graph - ConceptNet, providing a deep contextual backdrop against which causal relationships can be examined.

2. \textbf{Counterfactual Reasoning Enhancer (CRE)} introduces hypothetical `what-if' scenarios to test and refine the AI's causal inferences, ensuring that identified causal links are robust and not merely correlational.

3. \textbf{Context-Aware Prompting Mechanism (CAPM)} crafts tailored prompts that encapsulate enriched context and counterfactual insights, directing LLMs toward more precise and accurate causal reasoning.

The CARE-CA framework's unique strength lies in its seamless integration of structured knowledge from ConceptNet with the contextual understanding capabilities of LLMs. We extract relevant concepts and causal relationships from ConceptNet based on the input scenario. The extracted knowledge is transformed into natural language statements and embedded into the context provided to the LLM. Then, we generate counterfactual scenarios using ConceptNet information to encourage more robust causal reasoning. To illustrate, consider this example:

Input scenario: "After heavy rain, the streets were flooded."

The following steps are then undertaken:
\begin{enumerate}
    \item{ConceptNet extraction: Relevant concepts (rain, flood, street) and relationships (Rain CapableOf CauseFlooding) are extracted.}
    \item{Contextual embedding: We add context like "Rain is capable of causing flooding, especially in urban areas with poor drainage."}
    \item{Counterfactual enhancement: We introduce a counterfactual scenario such as "If the city had better drainage systems, would the streets still flood after heavy rain?"}
\end{enumerate}

This integrated approach combines structured, explicit causal knowledge from ConceptNet with the flexible, context-aware reasoning of LLMs, resulting in more robust and nuanced causal reasoning capabilities. We evaluate CARE-CA's performance using accuracy, recall, precision, and F1 scores in a single variable test.

To provide readers with additional context, we provide an example of a prompt for COPA Dataset below:\\\\
\fbox{
\begin{minipage}{\dimexpr 0.95\columnwidth\relax}
\textbf{Input:} ``Shadows are formed when a light source illuminates an object, creating a dark area on the opposite side. Given that `My body cast a shadow over the grass,' which hypothesis seems more plausible based on the understanding of shadows?\\

\textbf{Counterfactual statement:} ``If the grass were on fire, my shadow would have been the least of my concerns.''\\

\textbf{Hypothesis 1:} `The sun was rising.' (providing the light that cast the shadow)\\

\textbf{Hypothesis 2:} `The grass was cut.' (which is a condition unrelated to shadow formation)
    \end{minipage}
}\\\\

\subsection{Datasets}

To develop and evaluate our CARE-CA framework, we employ six distinct datasets. Each dataset serves a specific function within our research, ranging from training the model's causal reasoning capabilities to evaluating its performance in various causal reasoning tasks.
All experiments were performed with a dataset split of 75\%-25\% for train and test sets, and 3 runs were conducted for each dataset-model combination. We evaluated 5 LLMs - GPT-3.5, Mistral 7b, Gemini Pro, Llama 2, T5 using 5 datasets - COPA, Timetravel, CLadder, Com2sense and e-care, and then compared the above LLMs with our proposed method CARE-CA.

\textbf{Dataset(s) for Causal Relationship Identification (CRI):}
\begin{itemize}
    \item \textbf{CLadder and Com2Sense:} 
    \textit{Composition:} Derived from narrative texts, these datasets are crafted to pinpoint explicit causal links within a narrative context.\\
    \textit{Purpose:} They provide foundational training for the model's explicit causal reasoning abilities, allowing it to recognize and understand causal relationships within complex text structures.
\end{itemize}

\textbf{Dataset(s) for Counterfactual Reasoning (CR):}
\begin{itemize}
    \item \textbf{TimeTravel:} 
    \textit{Composition:} This dataset presents hypothetical scenarios that challenge the model to reason about events that did not occur.\\
    \textit{Purpose:} It is crucial for enhancing the model's counterfactual reasoning, teaching it to contemplate different possibilities and their implications.
\end{itemize}

\textbf{Dataset(s) for Causal Discovery:}
\begin{itemize}
    \item \textbf{COPA and e-care:} 
    \textit{Composition:} COPA focuses on scenarios that require understanding potential outcomes and alternate realities, while e-care contains medical narratives that add domain-specific intricacies.\\
    \textit{Purpose:} These datasets are utilized to challenge the model in discovering underlying causal mechanisms within varied and domain-specific contexts.
\end{itemize}

Each dataset contributes uniquely to the robustness of the CARE-CA framework, ensuring comprehensive coverage across the spectrum of causal reasoning tasks.

\subsection{Proposed Dataset}
We also propose a new dataset called CausalNet which is carefully designed to facilitate causal reasoning and counterfactual analysis research\footnote{\url{https://github.com/swagata15/causal-reasoning}}. Comprising 1000 carefully curated scenarios, this dataset presents a diverse set of causal and counterfactual questions, allowing researchers to explore the intricacies of cause-and-effect relationships in various contexts.

Each entry in CausalNet consists of the following components:

\textbf{Context:} A detailed narrative context provides the backdrop for each scenario. These narratives describe situations where multiple events or factors coincide, potentially influencing outcomes. The contexts are designed to be realistic and thought-provoking, setting the stage for causal reasoning and counterfactual exploration.

\textbf{Causal Questions:} For each scenario, a set of causal questions is provided to challenge the models' abilities in causal reasoning. These questions are categorized into two main types:

\textbf{Cause-Effect Questions:} These questions prompt models to identify less obvious factors that may have contributed to observed outcomes. Models must discern the subtle interplay of various events or conditions in determining the outcome.

\textbf{Counterfactual Questions: }Counterfactual questions explore how changes in the scenario's main cause might impact the outcome. Models are evaluated based on their capacity to predict the consequences of hypothetical alterations to the causal factor.

\textbf{Choices and Answers:} Each question is accompanied by a set of choices, one designated as the correct answer. For cause-effect questions, the choices represent potential influencing factors, while for counterfactual questions, the choices depict possible outcomes under different circumstances. The correct answers are carefully labeled to facilitate evaluation.

% The CausalNet dataset contributes to advancing natural language understanding and reasoning capabilities. It enables researchers to explore and enhance models' causal reasoning skills, paving the way for more interpretable and context-aware AI systems.

CausalNet was meticulously constructed using a multi-step process to ensure its quality and relevance.
\begin{enumerate}
\item{\textbf{Initial Generation:} We utilized GPT-4's advanced language capabilities to generate an initial set of 1,500 scenarios. Each scenario was designed to include a context, causal questions, and counterfactual questions.}

\item{\textbf{Prompt Engineering:}
% We crafted specialized prompts to guide GPT-4 in generating diverse, realistic scenarios across various domains (e.g., science, social interactions, economics). The prompts were iteratively refined to ensure the generation of high-quality, causally rich content.
We used the following prompt:
Develop a dataset composed of entries that challenge and enhance machine learning models' understanding of causal relationships and counterfactual reasoning across various domains. Each entry in the dataset should follow this structure:
``Context": A detailed description of a scenario that outlines a complex situation involving causal relationships.
``Questions": A set of questions focusing on (1) identifying causal effects within the context and (2) exploring counterfactual scenarios, with multiple-choice answers to infer the model's reasoning capabilities.}

\item{\textbf{Filtering and Refinement:} The initial set was filtered down to 1,000 high-quality scenarios. This process involved removing duplicates, overly simplistic scenarios, and those with ambiguous causal relationships. CausalNet is designed to bridge the gap between academic causal reasoning tasks and real-world applications. It covers a wide range of fields, mirroring the complexity of real-world causal reasoning tasks in areas such as healthcare, policy-making, and business decision-making. The richness of the dataset is ensured by adding many scenarios that require multi-step causal inference, simulating the complexity of real-world problem-solving. The dataset also incorporates subtle contextual cues that influence causal relationships, reflecting the nuanced nature of real-world causality. Scenarios often conclude with questions about potential interventions or decisions, aligning with practical applications of causal reasoning in fields like management and public policy. This tests LLMs on their practical decision-making ability.}

\item{\textbf{Verification Process:}
Due to the AI-generated nature of CausalNet, we employed a stringent human verification process to ensure that the dataset meets the highest academic standards. All authors of this research effort reviewed a subset of the scenarios for logical consistency and real-world relevance. Based on feedback, we iteratively refined the dataset, adjusting scenarios and questions to improve clarity and causal validity. We tested the refined dataset against existing causal reasoning benchmarks to ensure its uniqueness and added value to the field.}
\end{enumerate}

% \begin{figure*}[ht]
%   \centering
%   \includegraphics[angle=270, trim = 5.5cm 0cm 5.5cm 0cm, width=\linewidth]{model_acc_datasets.pdf} % Adjust the image width as needed
%   \caption{The graph demonstrates that the CausalNet dataset consistently improves the performance of all models, with T5 showing the most significant increase, reaching 94.2\% accuracy. This trend suggests that CausalNet's structure and content may be particularly effective in enhancing the causal reasoning capabilities of various language models, regardless of their architectural differences.}
%   \label{fig:image1}
% \end{figure*}

\section{Results}

The performance was quantitatively assessed through mean accuracy, precision, recall, and F1 scores which is illustrated in Figure \ref{fig:both-images}.

\subsection{Causal Discovery}
We examine CARE-CA's capability to unearth hidden or implicit causal relationships within complex scenarios. Our method showcased superior accuracy (76\%) on the COPA dataset, emphasizing the framework's strength in integrating contextual and counterfactual insights to uncover underlying causal mechanisms. Interestingly, GPT-3.5 and Gemini Pro also performed well, with accuracies of 73.3\% and 70.1\%, respectively, indicating their potential in learning causal patterns. The lower performance of models like XLM-RoBERTa and DeBERTa, with accuracies of 53.2\% and 51.8\%, respectively, could stem from their less effective handling of the dataset's counterfactual and causal scenarios without specific fine-tuning.

On the Ecare \cite{du2022ecare} dataset, our method also performed well with 85.9\% accuracy, compared to the next closest decoder model performance of T5 at 84\%.

\subsection{Causal Relationship Identification} 
The objective is to assess CARE-CA's proficiency in recognizing explicit causal links within narrative contexts. On the Cladder \cite{jin2023cladder} dataset, the CARE-CA model led with a standout performance, achieving a 63\% accuracy, indicating its strong capability to identify causal relationships.  The decoder model T5 highlighted its proficiency with a balanced performance, showcasing the effectiveness of its decoding capabilities in causal reasoning tasks.

On the Com2sense \cite{singh2021com2sense} dataset, the decoder models encountered diverse challenges, with CARE-CA again leading at 67.1\% accuracy, suggesting its consistent ability to navigate causal reasoning tasks. 

\begin{figure*}[htbp]
    \centering
    \begin{subfigure}[b]{0.49\textwidth}
        \centering
        \includegraphics[width=\textwidth]{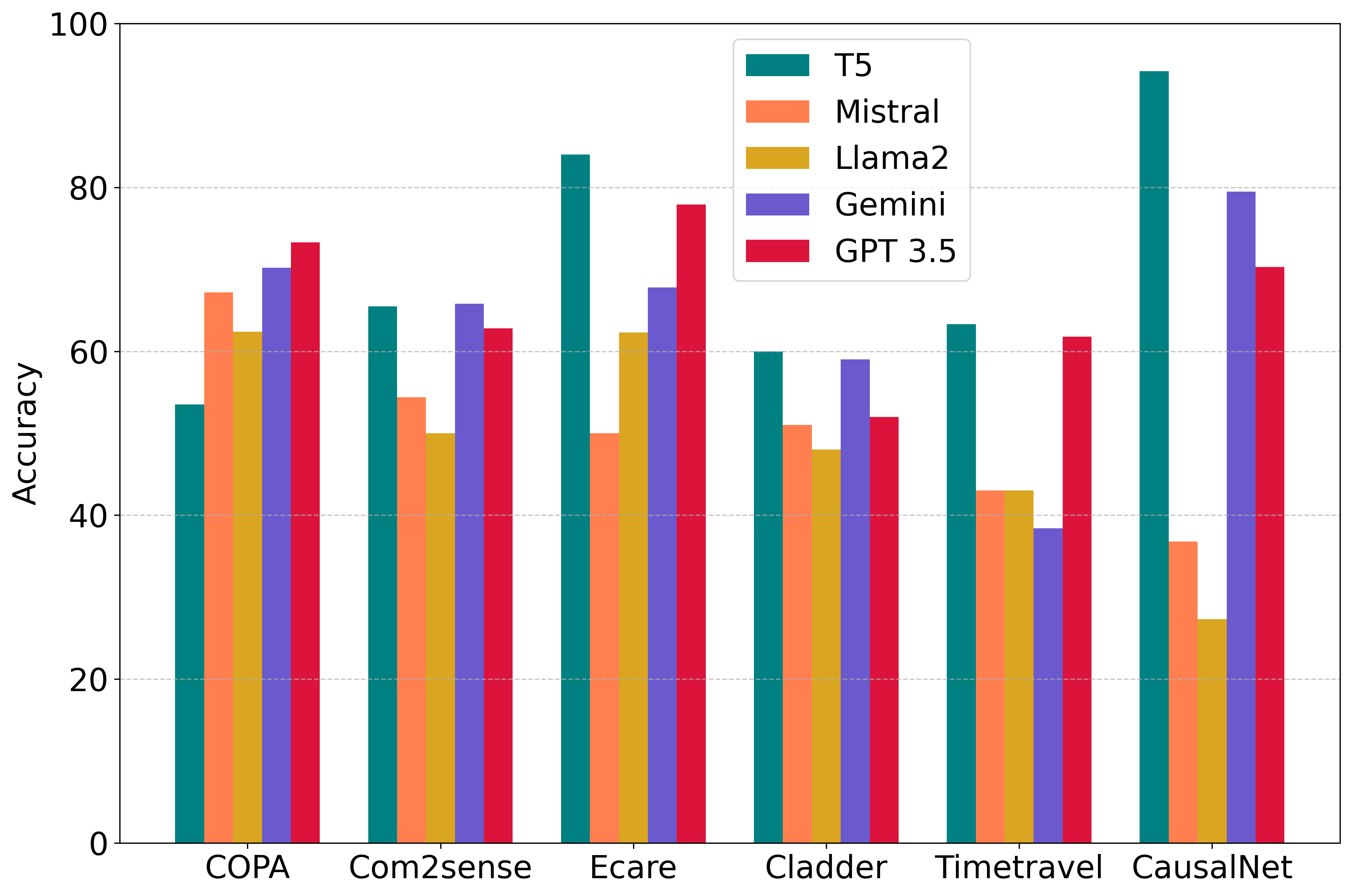}
        \caption{CausalNet dataset enhances performance across all models. T5 shows highest improvement with 94.2\% accuracy. Results suggest CausalNet's effectiveness in boosting causal reasoning capabilities.}
        \label{fig:image1}
    \end{subfigure}
    \hfill
    \begin{subfigure}[b]{0.49\textwidth}
        \centering
        \includegraphics[width=\textwidth]{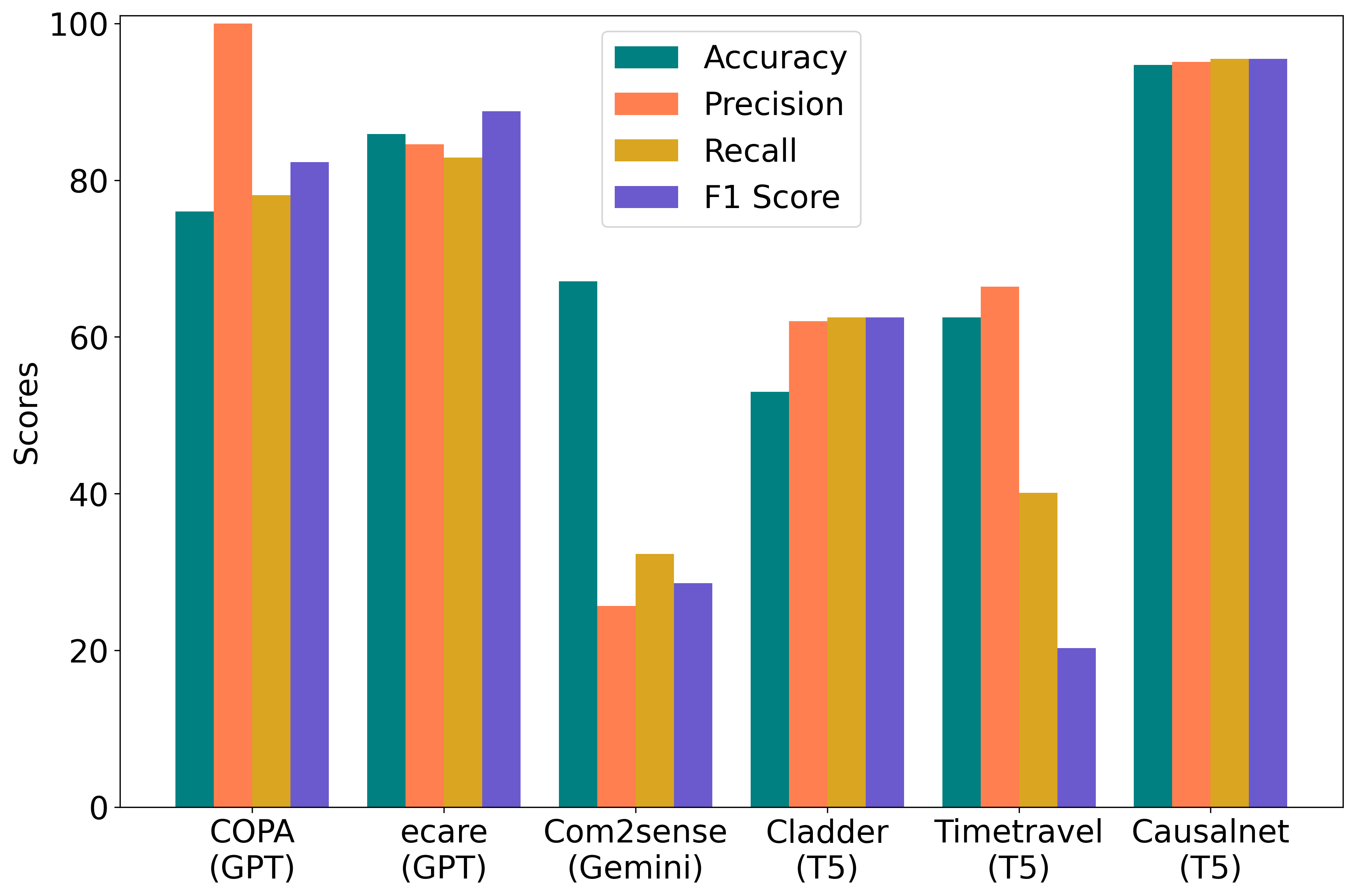}
        \caption{CARE-CA model excels in causal reasoning across datasets and tasks. On CausalNet, it achieves 94.6\% mean accuracy, demonstrating superior performance in diverse causal contexts.}
        \label{fig:image2}
    \end{subfigure}
    \caption{Performance comparison of causal reasoning models across datasets, highlighting the effectiveness of the CausalNet dataset and the CARE-CA model.}
    \label{fig:both-images}
\end{figure*}

On our CausalNet dataset, CARE-CA's remarkable accuracy of 94.6\% sets a high benchmark, emphasizing the model's superior causal reasoning capabilities. The T5 decoder model mirrored this high performance with a 94.2\% accuracy, showcasing the strength of decoder architectures in extracting and interpreting causal relationships from data. \ref{fig:image2} illustrates the performance of the CARE-CA model across multiple datasets. It is clear that the model demonstrates strong performance observed on CausalNet.

\subsection{Counterfactual Reasoning}
Here we test CARE-CA's ability to reason with hypothetical scenarios and their implications for understanding potential outcomes. The timetravel dataset \cite{qin-etal-2019-counterfactual}, focused on counterfactual reasoning, highlighted models' challenges in understanding hypothetical scenarios.  The Gemini Pro and Llama models scored 38.4\% and 24.2\%, respectively, suggesting that despite their extensive training data, they might struggle with tasks requiring deep counterfactual inference, underscoring the importance of specialized training or prompting for such tasks.T5 and GPT 3.5 models performed well with 61.7\% and 63.2\% respectively. Our method got a slight jump in accuracy from the best-performing decoders; however, due to information overload, it could not compete with relatively more straightforward encoders such as ALBERT with 68\% accuracy.

The CARE-CA framework demonstrated superior performance across various causal reasoning tasks compared to traditional LLMs. This exceptional performance (94.6\% accuracy) on our novel CausalNet dataset shows robustness in handling diverse causal reasoning tasks, effective integration of explicit knowledge and counterfactual reasoning in real-world scenarios, and the ability to generalize causal understanding across various contexts.

The hybrid approach of combining explicit knowledge (ConceptNet) with implicit reasoning (LLMs) creates a more comprehensive causal understanding while the inclusion of counterfactual reasoning allows for more robust causal inferences and hypothesis testing. CARE-CA's architecture enables better adaptation to different contextual nuances in causal scenarios. The framework effectively manages the trade-off between leveraging external knowledge and avoiding information overload.

\textbf{Remark:} The observed performances underscore the complexity of causal reasoning tasks and the varying abilities of models to address them. The CARE-CA framework's superior performance across several tasks suggests that its hybrid approach, which leverages explicit causal knowledge and counterfactual reasoning, significantly enhances causal inference capabilities. LLMs exhibit strong foundational abilities in causal reasoning, likely benefiting from their diverse pre-training. However, tasks requiring nuanced understanding or domain-specific knowledge, such as counterfactual reasoning and causal explanation, highlight the limitations of LLMs and the value of specialized training or frameworks like CARE-CA.

% \definecolor{highlight}{rgb}{1,0.75,0.75}

\begin{table*} 
\centering
\fontsize{10}{10}\selectfont  % This sets the font size to exactly 10pt with 10pt leading
\begin{adjustbox}{width=0.95\textwidth,totalheight=0.95\textheight} % Adjust table size to fit page
\begin{tabular}{lllllll}
\toprule
Experiment & Dataset & Model & Mean Accuracy & Mean F1 & Mean Precision & Mean Recall\\
\midrule
\multirow{12}{*}{Causal Discovery} 
 & \multirow{12}{*}{COPA} 

 & CARE-CA & 76.0 & 82.3 & 1.0 & 78.1\\
 && BERT & 69.2 & 66.3 & 70.0 & 68.6 \\
 && RoBERTa & 57.2 & 56.2 & 58.3 & 61.1 \\
 && XLM-RoBERTa & 53.2 & 47.0 & 52.1 & 56.2 \\
  && ALBERT & 62.2 & 63.1 & 64.0 & 66.2 \\
 && DeBERTa & 51.8 & 0.0 & 0.0 & 0.0 \\
 && Llama2 & 62.4 & 56.0 & 87.0 & 68.0 \\
 && T5 & 53.5 & 1.0 & 54.0 & 70.0 \\
 && Mistral & 67.2 & 67.2 & 1.0 & 87.1 \\
 && GPT-3.5 & 73.3 & 78 & 1.0 & 87.5 \\
 && Gemini Pro & 70.1 & 1.0 & 70.1 & 82.4 \\
\cmidrule{2-7}
 & \multirow{12}{*}{Ecare} 
 & CARE-CA & 85.9 & 88.8 & 84.6 & 82.9\\
 && BERT & 50 & 39.4 & 66 & 47.6 \\
 &&  RoBERTa & 49.7 & 51.5 & 50.8 & 73.1 \\
 && XLM-RoBERTa & 48.2 & 58.7 & 46.7 & 84.2 \\
 && ALBERT & 47.7 & 41.4 & 50.9 & 57.7 \\
 && DeBERTa & 46.6 & 63.6 & 46.6 & 100.0 \\
 && Llama2 & 62.2 & 60.0 & 63.8 & 56.7 \\
 && T5 & 84 & 84.8 & 80.5 & 89.6 \\
 && Mistral & 50 & 49.9 & 50 & 49.9 \\
 && GPT-3.5 & 77.8 & 75.9 & 83.3 & 69.7 \\
 && Gemini Pro & 67.8 & 63.0 & 74.4 & 54.5 \\
\midrule
\multirow{7}{*}{Counterfactual Reasoning} 
 & \multirow{7}{*}{Timetravel}
 & CARE-CA & 69.4 & 40.1 & 20.2 & 13.5\\
   && BERT & 56.3 & 6.0 & 11.0 & 5.0 \\
 && RoBERTa & 68.7 & 3.0 & 9.0 & 2.0 \\
 && XLM-RoBERTa & 56.9 & 5.0 & 10.0 & 3.0 \\
 && ALBERT & 68 & 6.0 & 11.2 & 4.0 \\
 && DeBERTa & 58.1 & 6.0 & 11.0 & 4.0 \\
 && Llama2 & 24.2 & 1.0 & 1.0 & 5.0 \\
 && T5 & 63.2 & 19.1 & 12.7 & 38.2 \\
 && Mistral & 27.5 & 2.0 & 1.0 & 6.0 \\
 && GPT 3.5 & 61.7 & 8.0 & 5.0 & 14.7 \\
 && Gemini Pro & 38.4 & 17.4 & 10.2 & 57.3 \\
\midrule
\multirow{10}{*}{Causal Reasoning Identification} 
 & \multirow{10}{*}{Cladder}
 & CARE-CA & 63.0 & 62.5 & 61.9 & 62.5\\
   && BERT & 53.0 & 48.6 & 52.3 & 52.4 \\
 && RoBERTa & 50.3 & 65.2 & 50.3 & 100.0 \\
 && XLM-RoBERTa & 49.5 & 64.3 & 49.5 & 99.3 \\
 && ALBERT & 49.4 & 46.2 & 40.5 & 68.9 \\
 && DeBERTa & 49.8 & 22.1 & 18.0 & 33.2 \\
 && Llama2 & 48.0 & 60.0 & 47.0 & 82.0 \\
 && T5 & 60.0 & 59.0 & 59.0 & 59.0 \\
 && Mistral & 51.0 & 59.0 & 52.0 & 70.0 \\
 && GPT 3.5 & 52.0 & 54.0 & 53.0 & 55.0 \\
 && Gemini Pro & 59.0 & 65.0 & 57.0 & 76.0 \\
\cmidrule{2-7}
 & \multirow{11}{*}{Com2sense}
 & CARE-CA & 67.1 & 28.6 & 25.7 & 32.3\\
   && BERT & 44.6 & 59.2 & 44.9 & 96.0 \\
 && RoBERTa & 45.5 & 1.0 & 3.0 & 1.0 \\
 && XLM-RoBERTa & 50.4 & 51.4 & 45.0 & 60.0 \\
 && ALBERT & 51.2 & 35.0 & 25.0 & 30.0 \\
 && DeBERTa & 45.3 & 60.0 & 45.6 & 96.5 \\
 && Llama2 & 50 & 20.0 & 10.0 & 13.3 \\
 && T5 & 65.4 & 63.4 & 46.2 & 53.4 \\
 && Mistral & 54.3 & 69.1 & 71.7 & 70.4 \\
 && GPT 3.5 & 62.8 & 23.2 & 30.4 & 28.0 \\
 && Gemini Pro & 65.8 & 25.2 & 31.6 & 28.0 \\
\cmidrule{2-7}
 & \multirow{11}{*}{CasualNet}
 & CARE-CA & 94.6 & 95.4 & 95 & 95.4\\
   && BERT & 39.0 & 21.8 & 15.2 & 39.0 \\
 && RoBERTa & 38.0 & 20.9 & 14.4 & 38.0 \\
 && XLM-RoBERTa & 37.5 & 20.4 & 14.9 & 37.5 \\
 && ALBERT & 33.8 & 19.3 & 27.2 & 33.8 \\
 && DeBERTa & 33.5 & 25.8 & 22.0 & 33.5 \\
 && Llama2 & 27.3 & 23.8 & 51.3 & 27.3\\
 && T5 & 94.2 & 94.5 & 95.0 & 94.2 \\
 && Mistral & 36.8 & 29.2 & 60.9 & 36.8 \\
 && GPT 3.5 & 70.3 & 70.9 & 84.6 & 70.3 \\
 && Gemini Pro & 79.5 & 80.0 & 83.8 & 79.5 \\
\bottomrule
\end{tabular}
\end{adjustbox}
\caption{The table summarizes performance metrics of encoders and decoders on three different tasks including causal discovery, counterfactual reasoning, and causal reasoning identification}
\label{tab:experimental_results}
\end{table*}

\section{Conclusion \& Future Work}

We present the CARE-CA framework as a significant advancement in enhancing the causal reasoning capabilities of large language models (LLMs). By integrating explicit knowledge from ConceptNet and employing counterfactual reasoning, CARE-CA bridges the gap between data-driven and knowledge-driven causal inference, offering a robust solution for various causal reasoning tasks. The evaluation on multiple datasets, including the newly introduced CausalNet, demonstrates that CARE-CA consistently outperforms traditional LLM approaches in accuracy, precision, recall, and F1 scores. This work not only contributes to the field of causal reasoning in AI but also paves the way for more interpretable and reliable AI systems. 
Our system works well under restrictive token constraints.\\
\textbf{Future Directions:}
These results pave the way for further research into hybrid models that combine the breadth of knowledge from resources like ConceptNet with the depth of understanding inherent in LLMs. Fine-tuning strategies, domain-specific model adaptations, and developing more comprehensive benchmarks like CausalNet are promising areas for future exploration.
Future research can further focus on expanding the multilingual capabilities of CARE-CA and further optimizing the framework to enhance its applicability across diverse domains and complex scenarios.

\section{Limitations}
In our research on the efficacy of causal reasoning in LLMs through the CARE-CA framework, we encountered several limitations that highlight areas for future exploration and improvement. Firstly, we were able to run CARE-CA only on best performing decoders of each dataset and compare the results. The comparison of CARE-CA on all decoders as well as on all encoders was a challenge due to computational resource constraints. Secondly, our focus on English limits the generalizability of our findings across languages and cultures; this opens a door for a need for multilingual datasets and cross-cultural validation. The challenge of applying our general causal reasoning framework effectively in domain-specific scenarios, such as those presented in the e-care dataset, indicates an opportunity for refining its adaptability to specialized fields. Additionally, the significant computational resources required by the CARE-CA framework may limit accessibility for those with constrained computational budgets, pointing to a need for optimization strategies. While CARE-CA enhances interpretability in causal reasoning tasks, further research is required to improve transparency and explain the model’s reasoning processes, especially for non-expert users. These limitations underscore the necessity for ongoing research to enhance the efficacy, inclusiveness, and applicability of causal reasoning models and invite the broader research community to address these challenges collaboratively.

\section{Ethics Statement}
Ethical considerations are paramount in research, particularly when LLMs are involved. We have strived to prevent the propagation of bias within CausalNet, the dataset we introduced in this work, by carefully curating and filtering the data to mitigate the inclusion of sensitive or discriminatory content. Furthermore, we have committed to transparency regarding the dataset's origins and potential implications, acknowledging the ethical responsibilities of conducting research with LLMs.

\bibliography{custom}

\begin{thebibliography}{24}
\providecommand{\natexlab}[1]{#1}

\bibitem[{Conneau et~al.(2019)Conneau, Khandelwal, Goyal, Chaudhary, Wenzek, Guzm{\'a}n, Grave, Ott, Zettlemoyer, and Stoyanov}]{conneau2019unsupervised}
Conneau, A.; Khandelwal, K.; Goyal, N.; Chaudhary, V.; Wenzek, G.; Guzm{\'a}n, F.; Grave, E.; Ott, M.; Zettlemoyer, L.; and Stoyanov, V. 2019.
\newblock Unsupervised cross-lingual representation learning at scale.
\newblock \emph{arXiv preprint arXiv:1911.02116}.

\bibitem[{Devlin et~al.(2018)Devlin, Chang, Lee, and Toutanova}]{devlin2018bert}
Devlin, J.; Chang, M.-W.; Lee, K.; and Toutanova, K. 2018.
\newblock Bert: Pre-training of deep bidirectional transformers for language understanding.
\newblock \emph{arXiv preprint arXiv:1810.04805}.

\bibitem[{Du et~al.(2022)Du, Ding, Xiong, Liu, and Qin}]{du2022ecare}
Du, L.; Ding, X.; Xiong, K.; Liu, T.; and Qin, B. 2022.
\newblock e-CARE: a New Dataset for Exploring Explainable Causal Reasoning.
\newblock Submitted on 12 May 2022, arXiv:2205.05849.

\bibitem[{He et~al.(2020)He, Liu, Gao, and Chen}]{he2020deberta}
He, P.; Liu, X.; Gao, J.; and Chen, W. 2020.
\newblock Deberta: Decoding-enhanced bert with disentangled attention.
\newblock \emph{arXiv preprint arXiv:2006.03654}.

\bibitem[{Jiang et~al.(2023)Jiang, Sablayrolles, Mensch, Bamford, Chaplot, Casas, Bressand, Lengyel, Lample, Saulnier et~al.}]{jiang2023mistral}
Jiang, A.~Q.; Sablayrolles, A.; Mensch, A.; Bamford, C.; Chaplot, D.~S.; Casas, D. d.~l.; Bressand, F.; Lengyel, G.; Lample, G.; Saulnier, L.; et~al. 2023.
\newblock Mistral 7B.
\newblock \emph{arXiv preprint arXiv:2310.06825}.

\bibitem[{Jin et~al.(2023{\natexlab{a}})Jin, Chen, Leeb, Gresele, Kamal, Lyu, Blin, Gonzalez~Adauto, Kleiman-Weiner, Sachan, and Sch{\"o}lkopf}]{jin2023cladder}
Jin, Z.; Chen, Y.; Leeb, F.; Gresele, L.; Kamal, O.; Lyu, Z.; Blin, K.; Gonzalez~Adauto, F.; Kleiman-Weiner, M.; Sachan, M.; and Sch{\"o}lkopf, B. 2023{\natexlab{a}}.
\newblock CLadder: Assessing Causal Reasoning in Language Models.
\newblock NeurIPS 2023; updated with CLadder dataset v1.5, arXiv:2312.04350.

\bibitem[{Jin et~al.(2023{\natexlab{b}})Jin, Liu, Lyu, Poff, Sachan, Mihalcea, Diab, and Sch{\"o}lkopf}]{jin2023can}
Jin, Z.; Liu, J.; Lyu, Z.; Poff, S.; Sachan, M.; Mihalcea, R.; Diab, M.; and Sch{\"o}lkopf, B. 2023{\natexlab{b}}.
\newblock Can large language models infer causation from correlation?
\newblock \emph{arXiv preprint arXiv:2306.05836}.

\bibitem[{Jin et~al.(2021)Jin, von K{\"u}gelgen, Ni, Vaidhya, Kaushal, Sachan, and Schoelkopf}]{jin2021causal}
Jin, Z.; von K{\"u}gelgen, J.; Ni, J.; Vaidhya, T.; Kaushal, A.; Sachan, M.; and Schoelkopf, B. 2021.
\newblock Causal direction of data collection matters: Implications of causal and anticausal learning for NLP.
\newblock \emph{arXiv preprint arXiv:2110.03618}.

\bibitem[{Kı{\c c}ıman et~al.(2023)Kı{\c c}ıman, Ness, Sharma, and Tan}]{kiciman2023causal}
Kı{\c c}ıman, E.; Ness, R.; Sharma, A.; and Tan, C. 2023.
\newblock Causal Reasoning and Large Language Models: Opening a New Frontier for Causality.
\newblock \emph{arXiv preprint arXiv:2305.00050}.

\bibitem[{Lan et~al.(2019)Lan, Chen, Goodman, Gimpel, Sharma, and Soricut}]{lan2019albert}
Lan, Z.; Chen, M.; Goodman, S.; Gimpel, K.; Sharma, P.; and Soricut, R. 2019.
\newblock Albert: A lite bert for self-supervised learning of language representations.
\newblock \emph{arXiv preprint arXiv:1909.11942}.

\bibitem[{Liu et~al.(2019)Liu, Ott, Goyal, Du, Joshi, Chen, Levy, Lewis, Zettlemoyer, and Stoyanov}]{liu2019roberta}
Liu, Y.; Ott, M.; Goyal, N.; Du, J.; Joshi, M.; Chen, D.; Levy, O.; Lewis, M.; Zettlemoyer, L.; and Stoyanov, V. 2019.
\newblock Roberta: A robustly optimized bert pretraining approach.
\newblock \emph{arXiv preprint arXiv:1907.11692}.

\bibitem[{OpenAI(2024)}]{gpt3.5}
OpenAI. 2024.
\newblock https://platform.openai.com/docs.

\bibitem[{Pe{\~n}a et~al.(2023)Pe{\~n}a, Morales, Fierrez, Serna, Ortega-Garcia, Puente, Cordova, and Cordova}]{pena2023leveraging}
Pe{\~n}a, A.; Morales, A.; Fierrez, J.; Serna, I.; Ortega-Garcia, J.; Puente, I.; Cordova, J.; and Cordova, G. 2023.
\newblock Leveraging Large Language Models for Topic Classification in the Domain of Public Affairs.
\newblock Accepted in ICDAR 2023 Workshop on Automatic Domain-Adapted and Personalized Document Analysis, arXiv:2306.02864.

\bibitem[{Qin et~al.(2019)Qin, Bosselut, Holtzman, Bhagavatula, Clark, and Choi}]{qin-etal-2019-counterfactual}
Qin, L.; Bosselut, A.; Holtzman, A.; Bhagavatula, C.; Clark, E.; and Choi, Y. 2019.
\newblock Counterfactual story reasoning and generation.
\newblock \emph{arXiv preprint arXiv:1909.04076}.

\bibitem[{Raffel et~al.(2020)Raffel, Shazeer, Roberts, Lee, Narang, Matena, Zhou, Li, and Liu}]{raffel2020exploring}
Raffel, C.; Shazeer, N.; Roberts, A.; Lee, K.; Narang, S.; Matena, M.; Zhou, Y.; Li, W.; and Liu, P.~J. 2020.
\newblock Exploring the limits of transfer learning with a unified text-to-text transformer.
\newblock \emph{The Journal of Machine Learning Research}, 21(1): 5485--5551.

\bibitem[{Romanou et~al.(2023)Romanou, Montariol, Paul, Laugier, Aberer, and Bosselut}]{romanou2023crab}
Romanou, A.; Montariol, S.; Paul, D.; Laugier, L.; Aberer, K.; and Bosselut, A. 2023.
\newblock Crab: Assessing the strength of causal relationships between real-world events.
\newblock \emph{arXiv preprint arXiv:2311.04284}.

\bibitem[{Singh et~al.(2021)Singh, Wen, Hou, Alipoormolabashi, Wu, Ma, and Peng}]{singh2021com2sense}
Singh, S.; Wen, N.; Hou, Y.; Alipoormolabashi, P.; Wu, T.-L.; Ma, X.; and Peng, N. 2021.
\newblock COM2SENSE: A Commonsense Reasoning Benchmark with Complementary Sentences.
\newblock In \emph{Findings of the Association for Computational Linguistics: ACL 2021}.
\newblock In Proceedings of Findings of the Association for Computational Linguistics: ACL 2021 (ACL-Findings). Contains 16 pages, 14 figures, and 11 tables.

\bibitem[{Speer, Chin, and Havasi(2017)}]{speer2017conceptnet}
Speer, R.; Chin, J.; and Havasi, C. 2017.
\newblock ConceptNet 5.5: An Open Multilingual Graph of General Knowledge.
\newblock \emph{AAAI Conference on Artificial Intelligence}, 4444--4451.

\bibitem[{Team et~al.(2023)Team, Anil, Borgeaud, Wu, Alayrac, Yu, Soricut, Schalkwyk, Dai, Hauth et~al.}]{team2023gemini}
Team, G.; Anil, R.; Borgeaud, S.; Wu, Y.; Alayrac, J.-B.; Yu, J.; Soricut, R.; Schalkwyk, J.; Dai, A.~M.; Hauth, A.; et~al. 2023.
\newblock Gemini: a family of highly capable multimodal models.
\newblock \emph{arXiv preprint arXiv:2312.11805}.

\bibitem[{Touvron et~al.(2023)Touvron, Martin, Stone, Albert, Almahairi, Babaei, Bashlykov, Batra, Bhargava, Bhosale et~al.}]{touvron2023llama}
Touvron, H.; Martin, L.; Stone, K.; Albert, P.; Almahairi, A.; Babaei, Y.; Bashlykov, N.; Batra, S.; Bhargava, P.; Bhosale, S.; et~al. 2023.
\newblock Llama 2: Open foundation and fine-tuned chat models.
\newblock \emph{arXiv preprint arXiv:2307.09288}.

\bibitem[{Weng et~al.(2023)Weng, Zhu, Xia, Li, He, Liu, Sun, Liu, and Zhao}]{weng2023large}
Weng, Y.; Zhu, M.; Xia, F.; Li, B.; He, S.; Liu, S.; Sun, B.; Liu, K.; and Zhao, J. 2023.
\newblock Large language models are better reasoners with self-verification.
\newblock In \emph{Findings of the Association for Computational Linguistics: EMNLP 2023}, 2550--2575.

\bibitem[{Ze{\v{c}}evi{\'c} et~al.(2023)Ze{\v{c}}evi{\'c}, Willig, Dhami, and Kersting}]{zevcevic2023causal}
Ze{\v{c}}evi{\'c}, M.; Willig, M.; Dhami, D.~S.; and Kersting, K. 2023.
\newblock Causal parrots: Large language models may talk causality but are not causal.
\newblock \emph{arXiv preprint arXiv:2308.13067}.

\bibitem[{Zhang et~al.(2023)Zhang, Bauer, Bennett, Gao, Gong, Hilmkil, Jennings, Ma, Minka, Pawlowski, and Vaughan}]{zhang2023understanding}
Zhang, C.; Bauer, S.; Bennett, P.; Gao, J.; Gong, W.; Hilmkil, A.; Jennings, J.; Ma, C.; Minka, T.; Pawlowski, N.; and Vaughan, J. 2023.
\newblock Understanding Causality with Large Language Models: Feasibility and Opportunities.
\newblock \emph{arXiv preprint arXiv:2304.05524}.

\bibitem[{Zhuang et~al.(2023)Zhuang, Chen, Ma, Li, Han, Qian, Bai, Feng, Zhang, and Liu}]{zhuang2023lens}
Zhuang, Z.; Chen, Q.; Ma, L.; Li, M.; Han, Y.; Qian, Y.; Bai, H.; Feng, Z.; Zhang, W.; and Liu, T. 2023.
\newblock Through the Lens of Core Competency: Survey on Evaluation of Large Language Models.
\newblock \emph{arXiv preprint arXiv:2308.07902}.

\end{thebibliography}

\end{document}